# A Heuristic Search Algorithm Using the Stability of Learning Algorithms in Certain Scenarios as the Fitness Function: An Artificial General Intelligence Engineering Approach

Zengkun Li

(Huawei 2012 Lab Central Software Institute Compiler and Programming Language Lab)

**Abstract:** This paper presents a non-manual design engineering method based on heuristic search algorithm to search for candidate agents in the solution space which formed by artificial intelligence agents modeled on the base of bionics. Compared with the artificial design method represented by meta-learning and the bionics method represented by the neural architecture chip, this method is more feasible for realizing artificial general intelligence, and it has a much better interaction with cognitive neuroscience; at the same time, the engineering method is based on the theoretical hypothesis that the final learning algorithm is stable in certain scenarios, and has generalization ability in various scenarios. The paper discusses the theory preliminarily and proposes the possible correlation between the theory and the fixed-point theorem in the field of mathematics. Limited by the author's knowledge level, this correlation is proposed only as a kind of conjecture.

## 1 Background

Although it is feasible to directly propose the engineering method in this paper, but in order to make it easier to explain the necessity and feasibility of this method and for readers without relevant background, it is necessary to introduce the background knowledge of the current mainstream artificial intelligence engineering method, and point out the challenges they face, as well as the reason why these challenges make them difficult to achieve artificial general intelligence.

Therefore, this paper introduces these methods only in the general process and methodological level without involving the details, and classify roughly in methodological means. This paper first introduces the current research status of cognitive neuroscience, and then discusses the two mainstream engineering methods: the artificial design method represented by meta-learning, and the biomimetic method represented by the neural architecture chip.

### 1.1 Status of cognitive neuroscience research

In order to clarify the causes of the limitations of current artificial intelligence engineering methods and their relationship with cognitive neuroscience, we must first introduce current research status of cognitive neuroscience.

At present, the research of cognitive neuroscience on the microstructure of human brain, mainly neurons and synapses, has been comparatively thorough, and drawn clear conclusions on their chemical characteristics and electrical properties. As for neurons, the spiking model is the most important characteristic. In terms of neuronal synapses, the features represented by STDP and LTDP are also fully studied.

However, research on macroscopic structure of the brain and its intelligent functionalities is still at a very early stage, and there is not any significant conclusion to the mechanism of human brain intelligence. Moreover, the main research methods in this area have also been questioned. The MRI-based method is one of the most important research methods in current cognitive neuroscience. In their paper, Eric Jonas and K.P. Cording discussed whether the MRI-based methods can be applied to analyze the principle of electronic chip[1], as a way of demonstrating that the methods cannot draw macroscopic functional structure conclusion from microscopic behavior characteristics.

The paper also reveals an opinion that cognitive neuroscience studies, without a programmatic theoretical guidance, will undoubtedly be difficult to study such a complex organ as the human brain. Therefore, while the cognitive neuroscience is carrying out the revelation of artificial intelligence, the cognitive neuroscience also urgently needs the results of artificial intelligence research to guide the direction of its research in turn. They have the relationship of mutually promoting.

In short, the current cognitive neuroscience cannot provide valuable inspiration for artificial intelligence engineering both at present and in foreseeable future. Similarly, cognitive neuroscience itself will continue to be in a state of lack of research orientation.

## 1.2 Artificial design methods

Artificial design methods is represented by meta-learning. Meta-learning is a broad concept, because it is in the early stages of development, there are many engineering methods, but there is only a few whose research object is really in line with the definition of the intelligent agent in this paper. Here are some of them: the gradient prediction based method[2], the Loss prediction based method [3], and the LSTM based updating method [4].

These methods are derived from deep learning, reinforcement learning and other traditional neural network learning algorithms. In these traditional learning algorithms, both the BP algorithm and the strategy gradient algorithm are independent of the neural network, and update the network connection weight outside the network, so it does not conform to the definition of the intelligent agent in this paper. Although the above-mentioned meta-learning methods take a similar approach to the traditional neural network learning method in modeling, but they incorporate the algorithm for updating the network connection weight into the network structure and take it as the learning target, so, the research object accords with the definition of intelligent agent in this paper.

However, these methods are completely dependent on artificial design. The human brain possesses tens of billions of neurons with an average of more than 100 connections between every two neurons. The current cognitive neuroscience mentioned above is still very elementary in terms of its conclusion drawn on the structure and the functional features of the brain. They basically cannot achieve a decisive inspiration for the realization of artificial general intelligence; at the same time, the human brain evolves over tens of thousands of years through natural selection. Even if it is not the optimal solution to general intelligence in human context, it is also an approximate optimal solution. It can be further deduced that, even if the implementation of artificial general intelligence is not because of the structure that is 100% identical with the human brain, at least the structure with approximate complexity. From the above we can see that, to achieve artificial general intelligence currently is equivalent to the case when you search for an approximate optimal solution in a vast space with very limited prior knowledge. If you rely on human exploration, it is difficult to imagine that you can find the right neural network structure and learning algorithm in the foreseeable future.

Of course, the speed of scientific exploration is not a linear process. In certain area, as scientific findings continue to emerge, the speed of exploration may show a quasi-exponential rise. However, the actual performance of the current intelligence generated by these methods is still far behind the human intelligence in terms of versatility and robustness. That is to say, the artificial design method is still in its initial extremely slow stage.

Another consequence of relying on artificial design is that it is impossible to interact with cognitive neuroscience and accelerate the development of each other. In the above method, both the micro-neuron characteristics and the macroscopic network structure, and even the network update algorithm, are far from the results of cognitive neuroscience research. The results of cognitive neuroscience research are incompatible with current approaches, and in turn, the results of current research methods cannot inspire cognitive neuroscience. Even if the artificial design method finally finds a correct path, what it achieved is probably just a logical equivalent of the human brain. Although this is not contrary to the goal of realizing artificial general intelligence, it is undoubtedly a detached detour.

## 1.3 Bionic methods

Bionics methods are represented by neural architecture chip like IBM's TrueNorth chip [5], which uses electronic components to partially simulate the biological characteristics of the human brain, such as the spiking model, neuron's plastic connection and so on. However, due to the same limitation of cognitive neuroscience's understanding of the human brain, these simulations are still at the micro level and are not involved in the macroscopic structure of the human brain's neural network. Even at the micro level, the simulation of the neuron and its connection cannot be proved to be complete. That is to say, some characteristics key to the generation of intelligence are likely to be missed.

Since the bionics methods adopt the loyal simulation of the biological characteristics of the human brain, there is naturally no issue for the artificial design methods, which can be said to be at the other extreme. However, the biggest problem of this method is that it cannot interact with cognitive neuroscience because it completely counts on the results of cognitive neuroscience research, thus the relationship between the two is one-sided, resulting in no possibility for mutual promotion and development.

## 2    Method of the paper

In view of the difficulties talked in 1.1 that currently are faced in cognitive neuroscience and artificial intelligence engineering, this paper presents a new artificial general intelligence engineering approach: the heuristic search algorithm using the stability of learning algorithm as the fitness function in certain scenarios. The following is a detailed introduction to the implementation of the algorithm.

## 2.1 Definition

Before introducing the methods proposed in this paper, two key concepts will be firstly defined.

### 2.1.1 The intelligent agent

The intelligent agent consists of two parts:

1. The material basis, on which the learning algorithm and reasoning algorithm perform simultaneously;

2. The learning algorithms, which act upon the material basis to change the property of it and thus its reasoning algorithms.

Although the mainstream artificial design methods, such as deep learning and reinforcement learning, are in line with the definition of artificial design methods, their learning algorithms are independent of their material basis - deep neural networks and thus they do not meet the first definition above. Therefore, from the perspective of the paper, these methods are not considered as the artificial intelligence engineering approach, which are excluded from the scope of discussion.

### 2.1.2 The stability of learning algorithm

The stability of learning algorithm is determined by two properties:

1. Learning algorithm, while performing the reasoning algorithm in the agent, will significantly change the property of material basis of the intelligent agents.

2. In the same scenario, the learning algorithm's change to the property of the material basis of the intelligent agents enables them to have a relatively consistent output.

The stability of learning algorithm is the core concept of this paper, and its definition even directly indicates the key ideas for the formation of the fitness function of the heuristic search algorithm proposed in this paper.

## 2.2 Model

The method in this paper is based on the heuristic search algorithm, which is a kind of optimization algorithm in the solution space to accelerate the search process for approximate optimal solutions, represented by genetic algorithm[6] which proposed by J. D. Bagley. In addition, the similar algorithms include particle swarm optimization algorithm, simulated annealing algorithm, ant colony optimization algorithm and so on.

Although all of them belong to same category of algorithms, different algorithms have their own scope of application. Among them, the simulated annealing algorithm and ant colony optimization algorithm can solve the unresolved problems by mapping them to the space-related problems and through the search on it, so it is more suitable for space-related problems. The method of the paper has its target of searching the artificial intelligence agents, whose problem characteristics are not naturally related to the space and therefore it is not suitable for these kinds of algorithms. And the genetic algorithm and particle swarm optimization algorithm solve problems by modeling them, and then evaluating the candidate solutions by fitness function. Therefore, they are fitter for the problems to be solved by the method of the paper. As for the use of whether genetic algorithm or particle swarm optimization algorithm or other similar ones, this paper does not restrict and believes that the specific choice should be based on engineering needs and experimental results. The following figure takes genetic algorithm as an example and describes the workflow of this kind of algorithm:

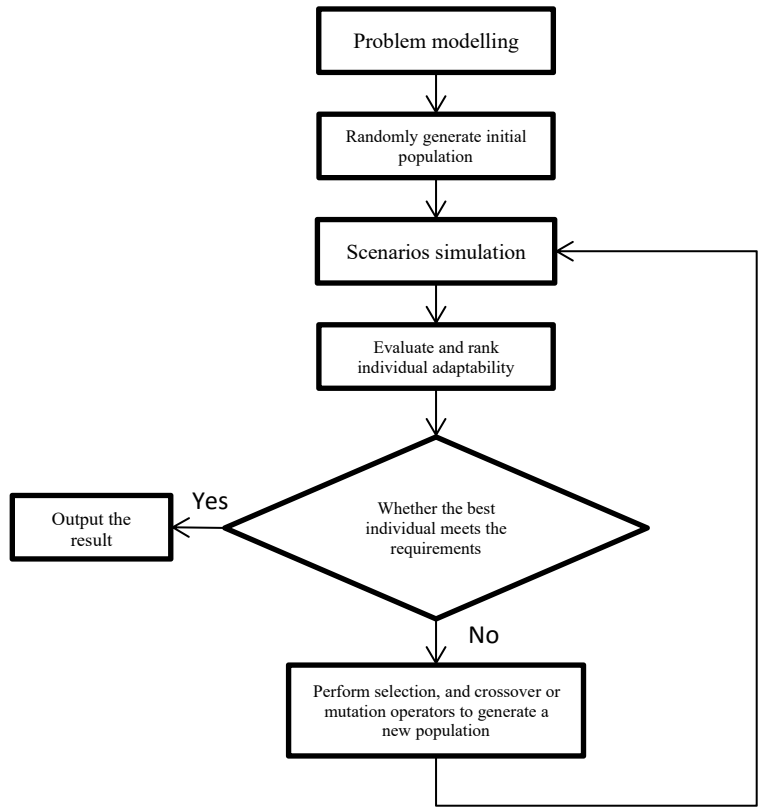

**Fig. 1 The operation flow of the genetic algorithms**

From the above, we can see that the construction of this kind of heuristic search algorithms mainly consists of three parts: problem modeling, defining of the fitness function and input of the fitness function – scenario simulation. The following will expound these three parts respectively.

### 2.2.1 Neural network

The method uses heuristic search algorithms to search for candidate agents in the solution space of artificial intelligence agents, so modeling the problem is to model the artificial intelligence agents. Taking into account the advantages of neural network in engineering realization and biological basis, this method uses neural network model to model artificial intelligent agents.

In the definition in 2.1.1, reasoning algorithms and learning algorithms depend on the material basis, which particularly refers to the two aspects in the neural network: corresponding network structure and network micro-characteristics. The network structure is reflected in the topological distribution of the connections between neurons and the distribution of the connection weights. If the connectionless ness is defined as the weight of 0, then the network structure can be simplified as the distribution of connection weights; the network micro-characteristics are reflected in the behavior of neurons, as well as changes in weight distribution of neuronal connection. From the current findings from cognitive neuroscience, changes in weight distributions of neuronal connectivity are mainly dependent on neuronal behaviors, which is of course still an open question, and as the research progresses further, different mechanisms may be introduced.

The current structure and micro-characteristics of neural network both determine the output for particular input, that is, the reasoning algorithm; and due to the existence of the key micro-characteristics, the structural changes of the neural network happened along with the reasoning process thus constitute a specific learning algorithm; as the network structure once changes, the output for the same input will also change, that is, the reasoning algorithm changes as the property of the material basis changes. In summary, this neural network model conforms to the definition of the intelligent agent in 2.1.1.

The following will be the respective introduction to the above-mentioned neural network structure and neural network micro-characteristics.

#### 2.2.1.1 Structure of the neural network

Because this method uses the heuristic search algorithm and completely abandons the artificial design so as to avoid the limitations of artificial design methods, the initial state of the intelligent agent must have the nature of covering the entire solution space, which particularly refers to the fully-connected structure in the neural network structure. Therefore, this method adopts the full connection between neurons as the initial state of the intelligent agent, and the initial weights of the connections are set at random values.

With the search algorithm progressing, part of the connections in the fully-connected structure is weakened or even canceled, while part of them are enhanced, which results in that the entire neural network gradually differentiates into various functional modules and evolves to the final form.

Concerning the specific implementation of fully-connected neural network, the author has implemented a prototype on the Nvidia GPU [7] platform, which, due to the layer-oriented design, is well suited for running on parallel computing frameworks such as GPUs, and the performances are proven to meet practical requirements.

Specifically, the fully-connected network is tiled in two tiers to form a classic Restricted Boltzmann Machine, but the difference is that one of the layers is an agent layer that does not have its own physical space but points to the other physical layer which is composed of real neuron in the network. Thus, the fully-connected structure between two layers of neurons is equivalent to the fully-connected structure inside the physical layer. The following figure shows the design of this structure:

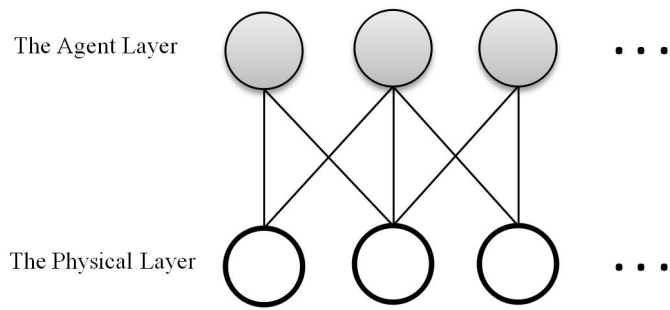

**Fig. 2 The hierarchical implementation of the fully-connected network structure**

The reason for the adoption of this kind of agent layer design is mainly for the simplification of the implementation and logical transformation from the fully-connected structure into the hierarchical structure. What needs to be mentioned is that the fully-connected structure will form the differentiation of functions in the later stage of algorithm and thus become sparse than its distribution in the initial state. However, this method runs the network in the same way regardless of the stage of the algorithm. This is because the network structure created during the search process is unpredictable, which is unlike the network structure that can be targeted for optimization in the artificial design methods.

Of course, this creates issues like the excessive computational burdens and the waste of computational power, which can be solved not only through the current rapidly growing computational power but also by the inspiration from cognitive neuroscience. Once cognitive neuroscience has made progress in the study of human brain structure, the results of which can be translated directly into the network structure design of the method, perhaps thus reducing the size of the fully-connected structure in the initial state or even directly regarding the sparse network structure as the initial state with then the application of targeted optimization. Of course, in the discussion of the current state of cognitive neuroscience research in 1.1, we can see that this kind of research on the macroscopic structure of human brain is still in its infancy, and the solutions to this problem cannot offer too much inspiration for a moment. The impact of the cognitive neuroscience study on this method is more reflected in the network micro-characteristics that are discussed in the next section.

#### 2.2.1.2 Micro-characteristics of the neural network

Micro-characteristics of the neural network are unlike the neural network structure discussed in 2.2.1 that can construct an initial state that covers the complete solution space on which to search. Therefore, the selection of the micro-characteristics of the network depends more on the prior knowledge of cognitive neuroscience. Fortunately, cognitive neuroscience has made great strides in the study of the microscopic properties of human brains, so this method is sufficient to get enough inspiration from the field of cognitive neuroscience, which will not be expounded again here because the paper has discussed it in 1.1.

It needs to be noted that this selection of micro-characteristics is fundamentally different from artificially designed neural network. As mentioned above, the micro-characteristics selected here are scientific facts that have been proved by cognitive neuroscience. However, artificially designed neural network is completely out of biological research and is basically an experience summary of engineering practice.

A key issue here is that while cognitive neuroscience offers a wealth of implications for artificial intelligence engineering in terms of human brain microscopic properties, some of the features in these conclusions may not have a functional effect on the production of intelligence, which might be a side effect of some kind of intelligent process without any key function; or just functional characteristics on physiologic dimension, such as the metabolic process of brain neurons; or even just an evolutionary legacy. Obviously, there is no effective way to identify this useless feature in advance, but the innate exploration mechanism of this method can solve this problem. Through a combination experiment of different features, it can identify to some extent which features perform better and then make the choice basing on it; or even a logical combination of a group of features such as one-hot coding, together with the network structure directly encoded into the solution space for heuristic search.

As for the details of implementation, should be analyzed case by case because there are so many characteristics of the human brain's nervous system. However, a key design must be particularly proposed here because it is the basis for realizing the characteristic of the spiking model which is almost sure to play a key role in the generation of intelligence.

In order to explain the necessity of this design, we need to give a brief introduction of the spiking neuron model. In the neuron system, each neuron sends electrical pulse signals to other neurons through neuronal synapses, and also receives electrical pulse signals from other neurons. It is the characteristic of electrical pulse signals from other neurons that determines how the neurons send electrical pulse signals. And put in the simplified model, the neurons will integrate their received electrical pulse signals, thereby generating their outward electrical pulse signals. These two processes often overlap in terms of time sequence, which can even lead to the more complex electrical spiking model. In order to simulate this feature, the neuron realization in this method adopts Ping pong Buffer mechanism whose implementation will be showed in the following figure:

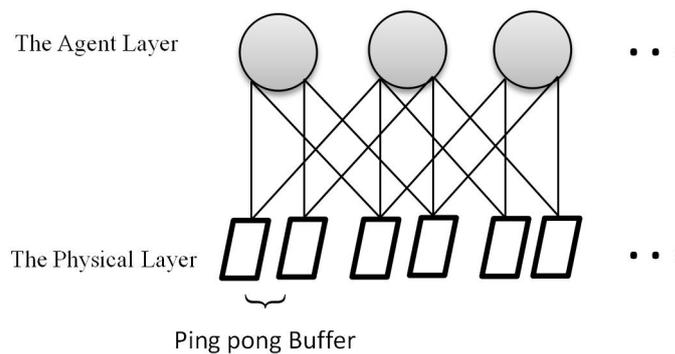

**Fig. 3 The implementation of Ping pong Buffer**

It can be seen in the figure above that there are two buffers in each simulated neuron. During a period of time, one of the buffers is responsible for integrating the input signals and preparation for sending out signal while the other buffer is responsible for simultaneously receiving the input signal. Entering into the next period of time, the data of the buffer that is responsible for the integration of input signals can be overwritten as it has completed the signal sending, and the buffer that is responsible for receiving the input signals in previous cycle also completed the storage of all data, so the role of two buffers will swap in the way that the buffer that is responsible for receiving the input signals in the previous cycle is ready to integrate and send off, and the other buffer that has completed the integration task starts to receive the input signals of the current cycle. In this way, the logic of time sequence in the real neuron system can be thus completely simulated in such a repeated process. This mechanism has been implemented in the prototype mentioned in 2.2.1 and works well.

### 2.2.2 Fitness function

### 2.2.2.1 Definition

Since the purpose of this method is to search for qualified intelligent agents in the solution space, then the goal is obviously to select those agents that can best perform the tasks in the scenarios. However, a truly qualified generic intelligent agents must not only be excellent in performing tasks within certain scenarios, but also have the capability to learn, adapt and complete different tasks in the generalization scenarios. Therefore, only taking the completion of tasks within the certain scenarios as the evaluation criteria cannot reach the goal.

In order to define a suitable fitness function to search for the above-mentioned eligible intelligent agents, here firstly puts forward the core hypothesis of this paper: if an agent can perform the same task repeatedly and stably in certain scenarios, the learning algorithm of that agent which changes the property of material basis of it then has the innate capability to generalize, thus making it adaptive to a series of other above-mentioned-task-related tasks. The learning algorithm is the one that is defined in 2.1.2 to have the stability under certain scenarios.

A popular expression of this assumption is to assume that there is already an intelligent agent that has reached our intelligence level as we require and has "learned" to complete the tasks in the current scenarios through learning. It should be noted here that the method of the paper differentiate from the traditional machine learning method in which the former one will directly search in the solution space for such an intelligent agent that has been directly in the state of "having learned" the tasks in current scenarios instead of the latter one that requires a certain training process to "learn" the current tasks. Then given such an intelligent agent that is capable of "learning" the current task and with a learning algorithm running on it, we have reason to assume that the agent is also capable of performing a series of tasks in other scenarios that are related to the current tasks, whose process of adaptation is just the learning and generalization.

Further, according to the definition in 2.1.1 that the intelligent agent contains the material basis and the learning algorithm and reasoning algorithm that run on it, as well as the requirements defined in item 1 in 2.1.2 for significant change of the property of the material basis, during the reasoning or behavioral process of the agents, the learning algorithm inevitably plays a role in changing the property of material basis and thus the reasoning algorithm. However, the above-mentioned assumptions require that the agent must have repeated and stable outputs under certain scenarios, which is equivalent to require the agent to have the relatively consistent reasoning algorithm and even material basis property. That is to say, in this case, the learning algorithm generates no significant change to the property of material basis. This seemingly contradiction is in fact not so, which can be explained by the fixed point theorem in mathematics, and it is precisely due to this nature that this learning algorithm possesses a natural generalization capability. A detailed discussion of this will be given in Section 3.

According to the above-mentioned assumptions, we can see that the key to defining the fitness function is to require the agent not only to complete the task, but also to stably repeat the task. Specifically speaking in reality, it is to allow the agent to perform the same task multiple times, and examine the results of its performance: the more the results of the performance accord with the expectation and more stable of these results, the higher the adaptability of the agent. Based on this theoretical assumption, we can clearly define a viable fitness function. The fitness function has two goals: the agent can behave as expected and the behavior can be repeated. In the actual implementation, it can be simplified to fetch the statistical average from the results of multiple execution.

It is worth mentioning that, logically, the stability of the learning algorithm is one of the important goals of the fitness function, but the ultimate goal of this paper is to search for agents. However, according to the definition in 2.1.1 and discussion of the modelling of intelligent agents in 2.2.1, it can be known that the learning algorithm in this paper completely depends on the material basis of the agent, that is, the agent itself. It can be said that finding out the learning algorithm that has the stability under certain scenarios leads to the candidate intelligent agent itself. This is the reason why this paper attaches so much importance to the fact that the learning algorithm runs on the material basis of the agent rather than it runs independently. This is not only an intrinsic feature of the biological intelligent agent, but also an important prerequisite for the feasibility of this paper.

As for the other goal of the fitness function, that is, the degree of task completion in certain scenarios, together with the characteristics of the specific scenario, jointly determines the intelligent level of the intelligent agent, which will be discussed in Section 2.2.3.

## 2.2.2.2 Mathematical conjecture

This section will elaborate on the relationship between the method of this paper and the fixed point of functions. Due to the limitation of the author's level, and the complexity of the fields covered in the method of this paper, here is the only the most coarsely presented discussion.

First of all, the learning algorithm and reasoning algorithm in Definition 2.1.1 are respectively expressed in the form of mathematical functions, expressing the learning algorithm as f and the reasoning algorithm as g. From Definition 2.1.1, we can see that the learning algorithm acts on the material basis and thus change g to form the adaptation to different scenarios. Therefore, f is a higher-order function whose domain and co-domain are both set G comprised of every function g.

At the same time, the goal of this paper is to make the agent perform the same tasks repeatedly and steadily in certain scenarios. In other words, it is that the reasoning algorithm g keeps stability under the action of f. Suppose that an agent is produced in the experiment, which can complete the task in a stable manner through n times of iterations. This is expressed in mathematical form as:

$$g_0 = f^n(g_0) \quad g_0 \in G \tag{1}$$

In this case, $g_0$ is a fixed point of the function f.

Because the goal of this paper is not only to find such a qualified agent, but also to hope that the agent can have the generalization ability and accomplish the tasks in scenarios that have not appeared in the experiment. The following is the discussion on the aspect of generalization.

Before going on to discuss generalization, further discussion of the reasoning algorithm set G is necessary. According to the definition in 2.1.1, reasoning algorithms are responsible for receiving input from the environment and producing output that form an agent's behavior. Assuming that all sets of environment input as set I and agent's output as set O, then there is $G: I \rightarrow O$; meanwhile, all the functions defined in a domain and co-domain can be defined as Cartesian product of the elements of the domain and co-domain, that is $G \Leftrightarrow I \times O$. Therefore, two tuples composed of element I and O can uniquely identify one reasoning algorithm $g \in G$. Assuming that the above corresponding two-tuples are $<i_0, o_0>$, the formula (1) can be rewritten as:

$$<i_0, o_0> = f^n(<i_0, o_0>) \quad i_0 \in I, o_0 \in O \tag{2}$$

Suppose there is another scenario that is out of the experiment, as well as the reasoning algorithm $g_1 \Leftrightarrow <i_1, o_1>$ that can complete the task. Due to the current reasoning algorithm of the agent is $g_0$, the agent's behavior can be represented as a two-tuple $<i_0', o_1>$ under such a scenario. Suppose that $<i_0', o_1>$ is in the neighborhood of points $<i_1, o_1>$, since $g_1$ is another fixed point of f and f will converge to this point in the neighborhood of it, then there is:

$$<i_1, o_1> = f^n(<i_0', o_1>) \quad i_1', i_1 \in I, o_1 \in O \tag{3}$$

Substituting the learning algorithm for the two-tuple by the above formula yields:

$$g_1 = f^n(g_0) \quad g_0, g_1 \in G \tag{4}$$

It can be seen that the iterative process of f, which is from $g_0$ to $g_1$ and even to all the fixed points, is the process of agent's gradual learning of the new scenarios and tasks, and this is called the generalization process.

Moreover, according to the uniqueness of function fix points, f must be a function set F, for arbitrary fix point $g_n$, $\exists f_n \in F$, makes the equation (4), so there must be a function $s: G \rightarrow F$, which always gives $f_n$ within the neighborhood of $g_n$.

To sum up, if the function $s$ exists, then the theoretical feasibility of the method in this paper will be guaranteed. And the essence of intelligence may also be revealed as the related properties of function $s$.

If the above hypothesis holds, then the generalization ability of the agent generated by this method is obviously related to the tasks in the experimental scenarios in terms of mathematics. In other words, different experimental scenarios will create different function $s$, resulting in different capabilities of the agent, which to some extent confirms the section 2.2.3 discussion on the scenario design.

Finally, it is worth mentioning that although in the above discussion a clear distinction is made between the learning algorithm f and the reasoning algorithm g, in practice both the learning algorithm and the reasoning algorithm operate on the same material basis, that is to say, the boundaries between these two may not be quite clear. Although, in the formulas discussed above, sometimes f receives g as a parameter, and receives the input-output two-tuple as a parameter in other occasions, the latter situation may be closer to the actual one. In the process of intellectual behavior, especially that in human beings, every inference may be accompanied by the process of relearning. This view can be well proved by the already existed evidence in the study of human memory that human memory must be re-stored at the same time while he or she reflects something. If it is expressed in the mathematical formula, it will lead to the last formula in this section:

$$G \circ s(G) \Leftrightarrow I \times O \tag{5}$$

### 2.2.3 Scenario simulation

Scenario simulation refers to the simulation of the external environment at which an intelligent agent is located and with which the agent interacts to complete a series of tasks. The agent's performance of these tasks becomes the input of the fitness function.

In engineering realization, there already have been many excellent achievements made in the field of reinforcement learning for reference. One of the most interesting new developments in it is the Unity ML-Agents [8], a Unity3D engine-based reinforcement learning platform introduced by Unity corporation. Although the specific engineering realization has a crucial impact on the quality and efficiency of the algorithm, it is not the key point of the method in this paper, so it will not be expounded further here. What is worthier of discussing is the relationship between the scenario design and the intelligence level of the final candidate agent. Then, some qualitative discussion relating to this aspect will be made.

The basic idea of the method in the paper is to search qualified candidate agents in certain scenarios, and then hope that the intelligent behaviors they acquire in certain scenarios can be generalized to other scenarios so as to achieve the purpose of general advanced intelligence. However, we can see from empirical fact that even the human intelligence has a limited ability of generalization. The same applies to the agents in this paper. It is hard to imagine that in a very simple scenario, an agent with complex intelligent behaviors can be generated. Therefore, it can be said that the complexity of the scenario basically determines the intelligent level of the final candidate agent.

Therefore, in order to generate an agent at a level comparable to that of human beings, the complexity of the scenario must be at least as high as the level of sophistication faced by early human intelligence. For example, in order to obtain food, early humans must carry out such tasks as hunting and collecting fruit. At the same time, they must also develop the viability of avoiding natural disasters and killing animals. From the perspective of intelligent evolution, the scenarios these early human faced are the important prerequisite for the formation of intelligence.

From this perspective, the ideas of the limited scenarios in the method of this paper are in fact very similar to those of human intelligence. The natural scenes that early humans were confronted with were rather limited. They did not come into contact with such complicated artifacts as modern humans did. However, it is under such a limited circumstance that humans have developed the general intelligence with a high degree of generalization. What is worth mentioning here is that although, on the surface, human intelligence has greatly improved compared to that of the early humans, the fact is that the human brain have not changed much compared with the early human brain in terms of the physiological level. The reason why human beings today create extremely rich of intelligent products is not due to the huge leaps in physiology but the result of cultural heritage. Therefore, it can be said that early humans, in fact, have basically possessed the physiological basis for producing advanced intelligence, but only being limited by the environment in which they live, leaving them to stay in the primitive phase of eating animal flesh raw and drinking its blood. In other words, the limited scenarios faced by early humans are sufficient for the formation of advanced intelligence, which provides some historical evidence of the validity of the methodological ideas in this paper.

However, this leads to another question, that is, if the above-mentioned hypothesis is true, then although the candidate agent generated by the method in this paper possesses the physical basis for generating advanced intelligence, due to the lack of the stage for the development of human culture and knowledge, its intelligent level is likely to remain at the level of early humans, who naturally cannot produce any real value to the present human.

There are two kinds of ideas to get rid of this dilemma. One of them is very naturally to construct a process similar to the development of human civilization. From historical experience, we can see that the process of human intellectual development is from figurative ones to abstract ones and from the special ones to the general ones. Although there is no definite conclusion, many studies show that human's language ability is an important foundation for abstracting abilities. If this conclusion is correct, then the target agent of the method in this paper must have at least linguistic competence before further intellectual development is possible. In the meantime, studies have shown that the emergence of human language is to better communicate and collaborate, and then the corresponding design of our scenarios may be joined by the tasks that require teamwork, so as to allow the agent to have the material basis for generating communication means, with the expectation of generating language-capable agents. Hunting for large animals is undoubtedly one of the most typical scenarios of this kind in the early days of human survival. When the agent possesses language skills, more advanced scenarios can be constructed so that the agent can continue to develop their intelligence, which even can be engaged by human beings so that the agent can learn the human cultures from the very beginning like another Homo sapiens, allowing its intelligent to become converge with that of human beings and eventually to serve human beings in a better way. Of course, linguistic competence, which is not necessarily the base for abstracting ability, may be the limited product of human's own ability, such as individual strength and memory ability. If the room for the memory of an artificial intelligence agent is large enough, it is not necessary to have a written record, and then we do not need to let the agent produce language behavior. For this is not the focus of this paper, so it will not be expounded further.

Of course, the ideas above may seem a little bit ridiculous. Even if the theory is feasible, the amount of calculation it takes far exceeds the existing capacity of mankind. Then the other idea, which is more viable, is to take a similar approach like that of the current machine learning: by constructing specific tasks for advanced scenarios with certain problem-orientation from the very beginning, which will empower agents to perform their advanced tasks so as to directly serve human beings. It is worth noting that the essential difference between this method and the traditional machine learning method is that although the agent generated by this method may not reach the level of general intelligence, its generalization ability is theoretically far beyond the methods for the time being. The agent will not only boast a strong generalization ability within its problem-oriented field, but even have the potential to evolve the ability to transfer intelligent behaviors across different domains. Therefore, it is still a valuable method.

From the second method in the above discussion, another point of view is introduced that the method of this paper is context-independent. In other words, the method is a universal one for acquiring generalized learning ability through certain scenarios. According to different scenario constructions, it can produce advanced agents in different fields, and even general ones that are comparable to the human beings. In term of this, there will be a more theoretical explanation after the introduction and connection to the fixed point theorem in Section 3.

With regard to the number of scenarios, it is also a vague concept because of the great uncertainty of the scenario design itself. A sufficiently complex scenario must be divided into multiple small scenarios, or during the process of performing the tasks, the intelligent agents will automatically switches the contexts, so the paper does not restrict the number of scenarios.

### 2.2.4 A summary of the model

The method of this paper is based on the heuristic search algorithm of problem-oriented modeling and adaptable evaluation, such as genetic algorithm or particle swarm optimization algorithm. Therefore, the overall flow is similar to that discussed in Section 2.2. In the aspect of modeling, this method adopts the fully-connected neural network model, plus the prior knowledge of the cognitive neurosciences about the micro-characteristics of the neural network to simulate the agent; in the aspect of the fitness function, the intelligent agent's repeated and stable performance of tasks under certain scenarios is mainly tested. Taking the same genetic algorithm as an example, the overall flow of the method of this paper will be illustrated below, in the form of marking the specific implementation of each step on the base of Figure 1:

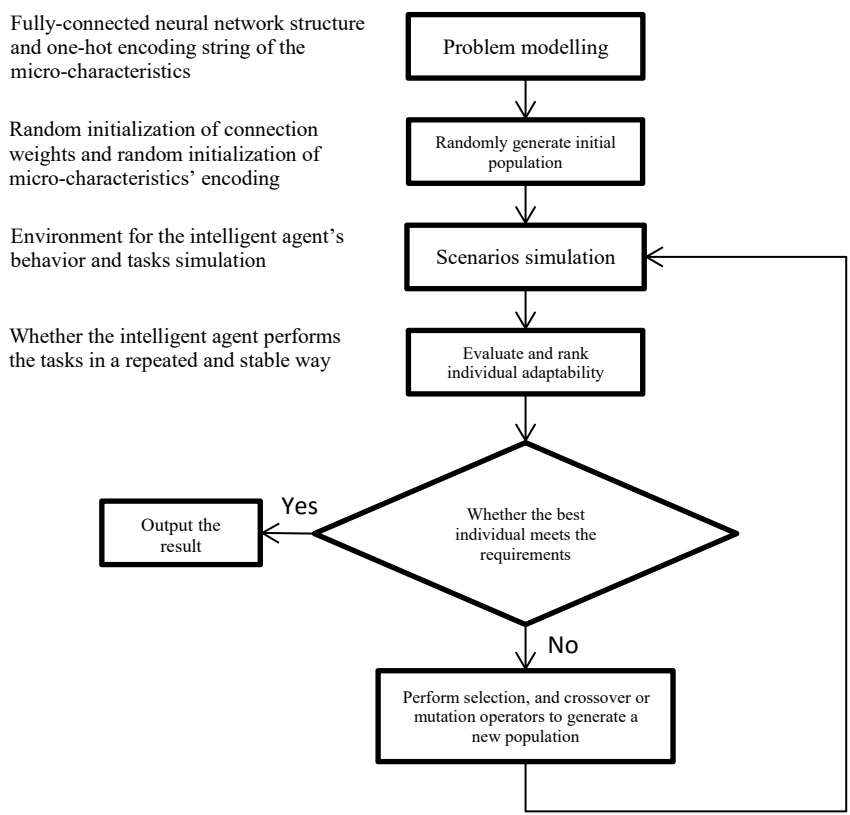

## 3 The optimization of the method

In engineering realization, there are many ways to optimize the method in this paper, such as using heterogeneous parallel computing framework to speed up the operation of neural network algorithm, tuning parameters of heuristic search algorithm or improving the algorithm itself so as to speed up its convergence speed, etc. However, the optimization of these specific realization can only improve the method of this paper to a certain extent, and what can really improve the efficiency of the algorithm in this paper comes from the research results of cognitive neuroscience.

From the previous discussion, we can see that the biggest challenge that this paper faces is the tremendous calculation amount and uncertainty caused by huge solution space for searching. Therefore, narrowing the solution space for searching is the only way to reduce the difficulty and raise the chance of success. At the same time, thanks to the modeling method of this paper, once the cognitive neuroscience has new and definite conclusions, they can directly provide reference for the modeling method of this paper, which can greatly narrow the solution space for searching in this paper. Therefore, drawing lessons from cognitive neuroscience research is the most important optimization method of this paper.

However, through the introduction of the current status of cognitive neuroscience research in section 1.1, it is clear that current cognitive neuroscience also needs to gain theoretical guidance from the field of artificial intelligence. Fortunately, the method of this paper is based on the current research results of cognitive neuroscience, which are enough to start the research. At the same time, the advantage of the method of this paper lies in its natural exploratory characteristics, which makes it possible to draw some enlightening results on cognitive neuroscience with limited prior knowledge. These results are reflected in the macroscopic structure of the neural network, and the functionality of micro-characteristics on the intelligence. Through experiments, some network structures beneficial to the agent will be discovered, and which kind of micro-characteristics has a significant effect on the agent performance may also be experimentally studied. These conclusions can be good guidance for further research in cognitive neuroscience, whose results will further narrow the search space of the method in this paper. As a result, a virtuous cycle of mutual promotion can be formed between these two, which can greatly accelerate the development of each other.

It is foreseeable that the method of this paper, in the early stage of its implementation, will also experience the same slower progress as that in the artificial intelligence engineering. However, as the experiment continues, its interaction with cognitive neuroscience will be more and more in-depth. It is believed that the pace of progress of both will be ushered in explosive growth.

In addition, the research on the basic theory of the mathematical model in the method of this paper can also provide theoretical guidance for it. Once the mathematical conjecture mentioned in Section 3 has been further studied or even confirmed and the mathematical theory can reveal the essence of intelligence, it is even possible to directly derive concrete methods of constructing artificial general intelligence.

In summary, in terms of the optimization of the method in this paper, while realizing the optimization of the basic algorithm, what is more important is that we should pay attention to interaction with cognitive neuroscience; and the study of theoretical basis is also a beneficial approach. It is believed that through these means, the method of this paper will receive enormous improvement in the pace of progress and rate of success.

## 4 Conclusion

The artificial general intelligence engineering method proposed in this paper, based on the heuristic search algorithm, completely abandons the artificial design methods and therefore avoids the shortcoming of slow progress in artificial exploration and low probability of realization; and its problem modeling only needs to rely on relatively more mature research results of cognitive neuroscience on the microstructure of neural system, and its field of exploration precisely lies in the aspects of research with more difficulty in cognitive neuroscience such as macro-structure of the neural system, intelligent generation mechanisms and so on. Therefore, the research results of this method can well guide the research of cognitive neuroscience. At the same time, the research results of cognitive neuroscience can in turn inspire the further exploration of modeling in this method. Hence, while this method is more viable in artificial general intelligence engineering, it can also create a better virtuous cycle with cognitive neuroscience and facilitate the accelerated growth of each other.

In addition, on the theoretical basis of this paper, the conjecture about the relativity with fixed point theorems is initially proposed. It can be seen from the related discussion that many properties of the fixed point theorem can be used to explain the viewpoints on the nature of intelligence proposed in this paper, which, to a certain extent, opens up new ideas for the theoretical basis research of the method in the paper.